\documentclass[journal]{IEEEtran}
\usepackage{graphicx}
\usepackage{booktabs}
\usepackage{multirow}
\usepackage{xcolor}
\usepackage{array}
\usepackage{amsmath}
\usepackage[colorlinks=true,linkcolor=blue]{hyperref}
\usepackage{subfigure}
\begin{document}
	
	\title{Prototype-based Cross-Modal Object Tracking}
	\author{Lei Liu, Chenglong Li, Futian Wang, Longfeng Shen, and Jin Tang
		\thanks{This research is partly supported by the National Natural Science Foundation of China (No. 62376004), the Natural Science Foundation of Anhui Province (No. 2208085J18), and the Natural Science Foundation of Anhui Higher Education Institution (No. 2022AH040014). (Corresponding author: Chenglong Li.)}
	}
	\markboth{Journal of \LaTeX\ Class Files,~Vol.~14, No.~8, August~2015}%
	{Shell \MakeLowercase{\textit{et al.}}: Bare Demo of IEEEtran.cls for IEEE Journals}
	
	\maketitle
	\begin{abstract}
		Cross-modal object tracking is an important research topic in the field of information fusion, and it aims to address imaging limitations in challenging scenarios by integrating switchable visible and near-infrared modalities. However, existing tracking methods face some difficulties in adapting to significant target appearance variations in the presence of modality switch. For instance, model update based tracking methods struggle to maintain stable tracking results during modality switching, leading to error accumulation and model drift. Template based tracking methods solely rely on the template information from first frame and/or last frame, which lacks sufficient representation ability and poses challenges in handling significant target appearance changes. To address this problem, we propose a prototype-based cross-modal object tracker called ProtoTrack, which introduces a novel prototype learning scheme to adapt to significant target appearance variations, for cross-modal object tracking. In particular, we design a multi-modal prototype to represent target information by multi-kind samples, including a fixed sample from the first frame and two representative samples from different modalities. Moreover, we develop a prototype generation algorithm based on two new modules to ensure the prototype representative in different challenges. The prototype evaluation module estimates the reliability of tracking results for each frame, determining whether to perform prototype extraction based on the tracking result. The prototype classification module predicts the modality state for each frame, facilitating the dynamic prototype updating of the associated modality samples. The multi-modal prototype forms a robust target representation under temporal variation and modality switch, and we integrate it into two tracking frameworks. Extensive experiments on the CMOTB dataset demonstrate the effectiveness and generalization of the proposed ProtoTrack against state-of-the-art methods.
	\end{abstract}
	
	
	\IEEEpeerreviewmaketitle
	
	\section{Introduction}\label{}
	
	Improving the adaptability of visual object tracking algorithms in challenging scenarios by fusing multiple sources of information is an important research area within the field of information fusion~\cite{liu2023visual, jiang2019hierarchical, tang2023exploring, zhang2020object}. Visual object tracking is a fundamental task in computer vision that involves predicting the state of a target object, including its position and size, in subsequent frames of a video sequence based on its initial state in the first frame. It plays a crucial role in a wide range of applications, including video surveillance~\cite{mishra2016study}, intelligent transportation~\cite{liu2023efficient}, and autonomous driving~\cite{gao2019manifold}. However, traditional visual object trackers often encounter difficulties in achieving robust tracking performance due to imaging limitation, making reliable tracking in all lighting and weather conditions a challenging task.
	
	To overcome this problem, many researchers are exploring the integration of multiple modalities, such as depth~\cite{Song13iccv, jiang2019hierarchical} or thermal infrared~\cite{li2016learning, li2017weighted, li2019rgb, li2021lasher, zhang2022visible, tang2023exploring, zhang2020object} data, to enhance tracking robustness in challenging scenarios. Incorporating multiple modalities can improve tracking performance by providing complementary information, but it also introduces new challenges. For example, depth sensors usually have limited range capabilities, typically up to 4-5 meters at most~\cite{Song13iccv}. Similarly, RGB-thermal sensors require accurate pixel-level alignment~\cite{li2016learning,li2019rgb,li2021lasher}, which can be challenging to achieve. The need for precise spatial registration and temporal alignment when using multiple modalities introduces complex and time-consuming pre-processes. These challenges make it harder to create high-quality multi-modal datasets, which are crucial for achieving effective tracking.
	
	In recent years, near-infrared (NIR) imaging has emerged as an important component in many surveillance cameras. NIR imaging allows switching between RGB and NIR modalities based on light intensity, providing an alternative approach to address the limitations of traditional multi-modal imaging platforms. Consequently, RGB-NIR cross-modal object tracking~\cite{li2022cross} has gained attention in the computer vision community. This approach leverages the strengths of both RGB and NIR imaging, making it easier to achieve robust tracking in diverse lighting conditions without the need for complex procedures involving precise spatial registration or temporal alignment. 
	
	Despite the recent advancements in visual object tracking, existing tracking methods still encounter challenges when it comes to adapting to significant target appearance variations during modality switch in cross-modal object tracking. For example, model update based tracking methods~\cite{nam2016learning, dai2020high, jung2018real, danelljan2019atom, bhat2019learning, danelljan2020probabilistic, mayer2022transforming} face difficulties in maintaining stable tracking results during modality switching. This struggle often leads to error accumulation and model drift, which ultimately degrade the tracking performance. Similarly, template based tracking methods rely solely on the template information from the first frame~\cite{bertinetto2016fully, wang2019fast, li2019siamrpn++, chen2020siamese} and/or the last frame~\cite{zhang2019learning, wang2021transformer, yang2018learning}. However, this approach lacks the sufficient representation ability to effectively handle significant changes in the target appearance. Consequently, template based methods encounter challenges in accurately addressing substantial target appearance variations during modality switches.
	
	Prototype learning has emerged as a promising approach for handling variations in target appearance and has shown great potential in various computer vision tasks~\cite{yang2018robust, li2021adaptive, yang2021part, liu2022intermediate, lu2023breaking, cheng2023prior}. By capturing the prototype representation of the target object, prototype-based methods can cope with diverse appearance in target object with different scenes. However, the application of prototype learning in cross-modal object tracking, specifically during modality switches, remains relatively unexplored. To address this problem, we propose a prototype-based cross-modal object tracker called ProtoTrack, which introduces a novel prototype learning scheme to adapt to significant target appearance variations, for cross-modal object tracking. 
	
	In particular, we design a multi-modal prototype to represent target information by multi-kind samples, including a fixed sample from the first frame and two dynamically updated representative samples from different modalities during online tracking. The fixed sample serves as a stable reference for tracking, providing an initial prototype representation of the target appearance. Besides, we utilize two dynamically updated samples to adapt to changes in the target appearance over time and across different modalities, providing adaptive prototype representations of the target appearance. By leveraging both the fixed sample and the dynamically updated samples, our approach effectively combines the strengths of both prototype representations, leading to more accurate and robust tracking performance across various modalities. Furthermore, our method ensures reliable prototype representations are available in any modality by considering multi-kind samples, thereby enhancing the stability of the cross-modal tracker.
	
	Moreover, we develop a prototype generation algorithm based on two new modules to ensure the prototype representative in different challenges, including the prototype evaluation module and the prototype classification module. The prototype  evaluation module estimates the reliability of tracking results for each frame, determining whether to perform prototype extraction based on the tracking result. The prototype classification module predicts the modality state for each frame, facilitating the dynamic prototype updating of the associated modality samples. In particular, to adapt to temporal changes in target appearance, we periodically update the multi-kind samples. Additionally, to adapt to modality switches in target appearance, we update the multi-kind samples of the corresponding modality when there is a change in the modality state. In this way, the multi-modal prototype forms a robust target representation under temporal variation and modality switch, and we integrate it into two tracking frameworks, including Stark \cite{yan2021learning} and OSTrack~\cite{ye2022joint}, for validating its generalization. 
	The contributions of this paper are summarized as follows:
	\begin{itemize}
		\item We propose a first prototype-based cross-modal object tracker called ProtoTrack to address significant target appearance variations in cross-modal object tracking.
		
		\item We design a multi-modal prototype that combines fixed and dynamically updated representative samples, enhancing tracking performance across various modalities.
		
		\item We introduce a prototype generation algorithm based on confidence evaluation and prototype classification modules to ensure the prototype representative in different challenges.
		
		\item We integrate the multi-modal prototype into two tracking frameworks and demonstrate its effectiveness and generalization through experiments on the CMOTB dataset.
		
	\end{itemize}
	
	\section{Related Work}
	In this section, we will provide a brief overview of the relevant research, focusing on three areas: visual object tracking, cross-modal object tracking and prototype learning. 
	\subsection{Visual Object Tracking}
	Visual object tracking is a challenging task that requires accurately localizing and tracking a target in a video sequence~\cite{xu2023learning,liu2023visual}. One critical factor that impacts the robustness of visual object trackers is target appearance variation. To address this issue, there are currently two main approaches.
	
	The first approach is model update based tracking methods. These methods update the model parameters using the backpropagation algorithm by collecting historical information during the online tracking. Examples of model update based tracking methods include MDNet~\cite{nam2016learning} and LTMU~\cite{dai2020high}. MDNet~\cite{nam2016learning} fine-tunes the classifier parameters by collecting features from positive and negative samples during tracking stage, effectively adapting to variations in target appearance. However, this method may lack robustness as it solely considers appearance representation. On the other hand, LTMU~\cite{dai2020high} integrates various cues, such as geometric, discriminative, and appearance cues, to generate sequential information, which is then fed into a cascaded LSTM to construct a meta-updater. This meta-updater outputs a binary score indicating whether the tracker should update its parameters based on the current frame information. This strategy significantly improves the decision accuracy for parameter updates in each frame. 
	While online gradient updates are involved in updating tracker parameters, they introduce two problems. Firstly, backpropagation, the strategy for online gradient updates, is not suitable for edge devices~\cite{yan2021learning}. Additionally, gradient updates incur time costs, which affect the inference speed. Secondly, in cross-modal tracking tasks where the target appearance varies across modalities, gradient updates become more unstable~\cite{li2022cross}, compromising the overall robustness of the tracker. 
	
	The second approach is template-based tracking methods. These methods update the target template through template collection and generation strategies. Examples of template-based tracking methods include UpdateNet~\cite{zhang2019learning} and Stark~\cite{yan2021learning}.  UpdateNet~\cite{zhang2019learning} predicts the accumulated template for the current frame using the initial template from the first frame, the accumulated template from the previous frame, and tracking result from the last frame. This method significantly enhances the tracker robustness without significantly affecting the inference speed. Stark~\cite{yan2021learning} explicitly models spatio-temporal information using transformer-inspired techniques~\cite{vaswani2017attention}. By aggregating the initial target template, current image, and dynamically updated template, Stark generates more discriminative spatio-temporal features for precise target localization. While Stark is effective in adapting to dynamic appearance differences over time, it is essential to also consider the variations in target appearance across modalities in cross-modal tracking scenarios.
	
	\subsection{Cross-modal Object Tracking}
	Changes in imaging modality can pose a significant challenge to cross-modal object tracking, as the appearance of the target may vary significantly between modalities. This variation can have a profound impact on the tracking performance, making it difficult to accurately localize and track the target over time.
	Li et al.~\cite{li2022cross} introduce the first benchmark for cross-modal object tracking. This benchmark contains 644 sequences and covers a wide range of real-world scenarios, providing a robust evaluation platform for cross-modal tracking algorithms.
	In addition to the benchmark, Li et al.~\cite{li2022cross} propose a modality-aware feature learning approach to mitigate the effects of cross-modal target appearance variations. However, one limitation of this method is the absence of explicit modeling of relationships between different time frames and modalities. As a result, effectively integrating information from both modalities and maintaining consistent tracking performance over time becomes more challenging.
	In contrast, our approach takes a different perspective by incorporating both temporal and modal information as a unified entity. We leverage multi-modal prototype to effectively capture complex patterns and dependencies between different time frames and modalities. In this way, our method achieves robust tracking performance across any modality, overcoming the limitations of previous approaches.
	
	\subsection{Prototype Learning}
	Prototype learning is a commonly used learning approach that is typically employed to learn a set of prototype representations for each category. These prototype representations are representative and can capture the information of the respective category. Prototype learning has been widely applied in various visual tasks, such as few-shot semantic segmentation~\cite{liu2022intermediate, wang2019panet, liu2020crnet, zhang2019canet, yang2021part}, few-shot object detection\cite{yan2019meta, lu2023breaking}, and person re-identification\cite{schumann2017deep, rao2023transg, tan2022dynamic, li2021diverse}.
	
	In the task of few-shot semantic segmentation, prototype learning usually involves learning a set of prototype representations from support images, which are then used as classifiers for segmenting query images. To improve the representativeness of the learned prototype representations, researchers have proposed a series of improved prototype extraction networks, such as feature alignment~\cite{wang2019panet}, cross-reference between support and query branches~\cite{liu2020crnet}, iterative mask refinement~\cite{zhang2019canet}, and semantic decomposition-and-match~\cite{yang2021part}.
	Similarly, in the task of few-shot object detection, it is common to learn a set of prototype representations from support images and apply these prototypes to the region features of query images for target detection~\cite{yan2019meta}. To enhance the diversity of the learned prototype representations, Lu et al. ~\cite{lu2023breaking} generate specific prototypes for each query image, significantly improving the quality of prototypes.
	Prototype learning also finds extensive application in person re-identification. For instance, Schumann et al. ~\cite{schumann2017deep} employ prototype learning to mitigate domain discrepancies by learning prototype representations in different domains, Li et al. ~\cite{li2021diverse} use prototype learning to decouple fine-grained body part representations, Tian et al. ~\cite{tan2022dynamic} utilize historical prototypes to better address occlusion challenges.
	
	Based on the above analysis, it is evident that prototype learning has been widely applied in the field of few-shot learning. As visual object tracking represents a typical one-shot local detection task, exploring the application of prototype learning in object tracking holds great potential, especially in mitigating cross-modal appearance variations through learned representative prototype representations.
	
	
	\begin{figure}[t]
		\centering
		\includegraphics[width=\linewidth]{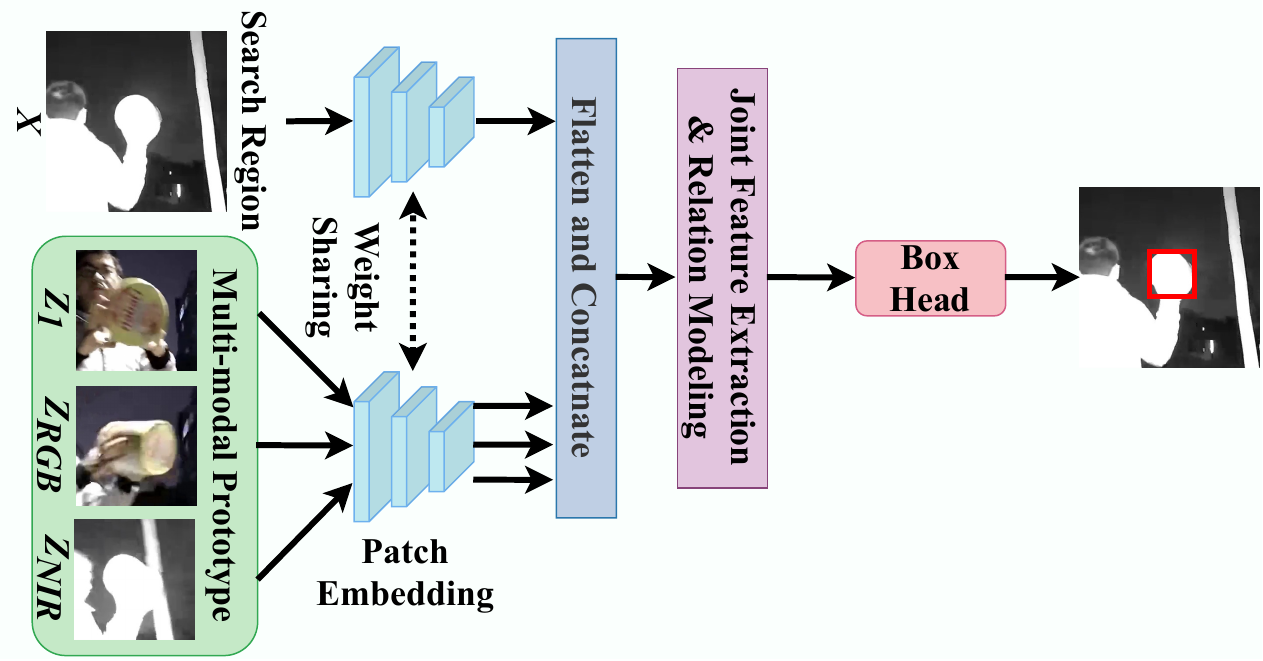}
		\caption{
			Framework for prototype-based cross-modal tracker.}
		\label{fig::overal_network}
	\end{figure}
	
	\section{Proposed ProtoTrack}
	In this section, we introduce ProtoTrack, the proposed prototype-based cross-modal object tracker that utilizes a novel prototype generation scheme to effectively adapt to significant target appearance variations in cross-modal object tracking.
	For clarity, we first introduce the tracking process of a cross-modal tracker utilizing pre-defined multi-modal prototypes. Next, we describe the prototype generation algorithm, which entails the extraction and updating of multi-modal prototype in the prototype-based cross-modal tracker. To control the updating of the multi-modal prototypes and ensure their representation in different challenges, we introduce the prototype evaluation module and prototype classification module.

	\subsection{Tracking Framework}
	In this work, we propose a prototype-based cross-modal object tracker designed to handle substantial variations in target appearance within cross-modal tracking scenarios. The network architecture is illustrated in Figure \ref{fig::overal_network}.
	\subsubsection{Overview}
	Our ProtoTrack follows the settings of state-of-the-art one-stream tracking paradigms~\cite{ye2022joint, yan2021learning, cui2022mixformer}. 
	Specifically, ProtoTrack takes the search region $X$ of the current frame and a multi-modal prototype as input. The multi-modal prototype represents target information using multi-kind samples, including a fixed sample $Z_1$ from the first frame, and two representative samples, i.e., $Z_{RGB}$ and $Z_{NIR}$, from different modalities during online tracking.
	Firstly, we partition the input image samples, i.e., $X$, $Z_1$, $Z_{RGB}$, $Z_{NIR}$, into 16x16 image patches and perform patch embedding~\cite{ye2022joint}, yielding encoded features, i.e., $H_X$, $H_{Z_1}$, $H_{Z_{RGB}}$, $H_{Z_{NIR}}$. Subsequently, these features are flatten and concatenated along the spatial dimension to form $H=[H_X;H_{Z_1};H_{Z_{RGB}};H_{Z_{NIR}}]$, which is then fed into a Vision Transformer(ViT)~\cite{dosovitskiy2020vit} for joint feature extraction and relational modeling. Finally, the output is further processed by a box head~\cite{yan2021learning} to predict the tracking results. Apart from the multi-modal prototype, the tracking pipeline of ProTrack aligns with existing one-stream tracking paradigms. For a more detailed description of the tracking pipeline, please refer to ~\cite{ye2022joint, yan2021learning, cui2022mixformer}.
	\begin{figure}[t]
		\centering
		\includegraphics[trim= 15mm 2mm 23mm 17mm, clip, width=\linewidth]{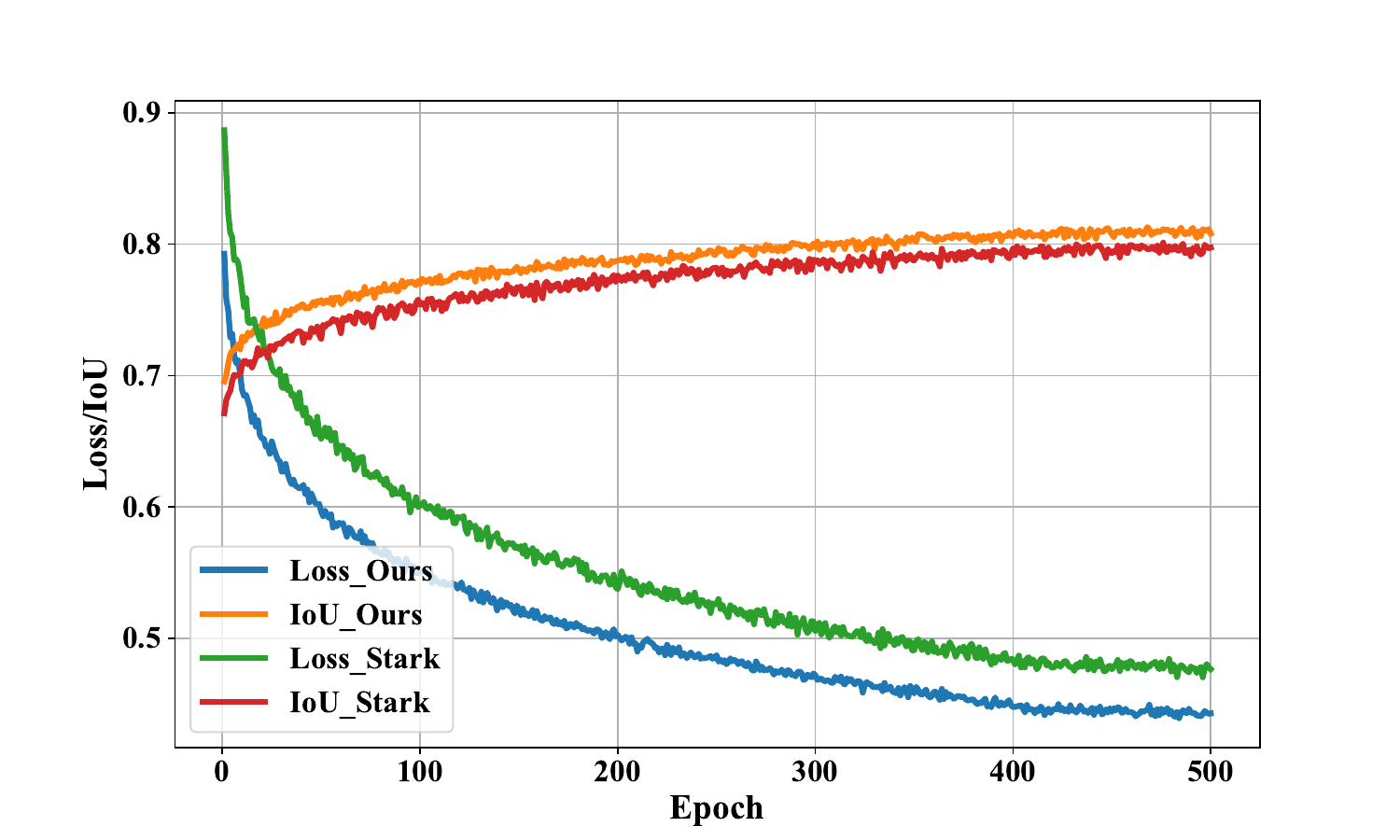}
		\caption{
			Comparison of training loss and accuracy on the CMOTB dataset~\cite{li2022cross} using the proposed multi-modal prototype and considering only temporal appearance variations in Stark~\cite{yan2021learning}.
		}
		\label{fig::lossiou}
	\end{figure}
	
	\begin{figure*}[t]
		\centering
		\includegraphics[width=\linewidth]{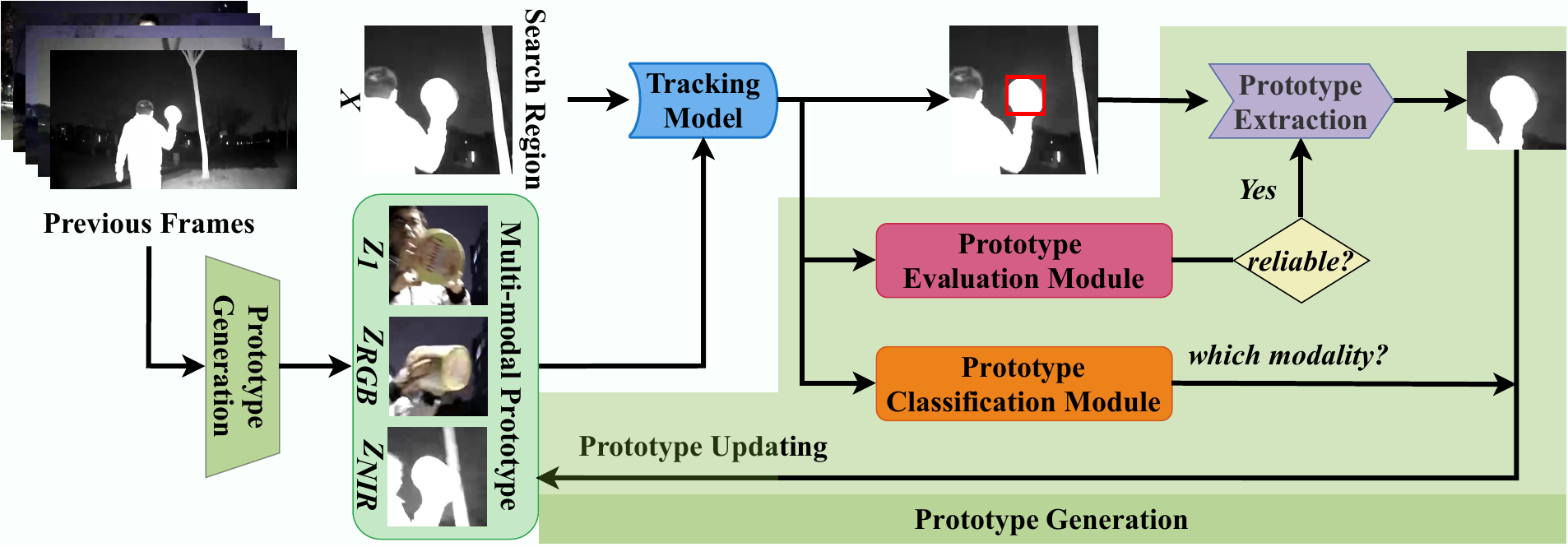}
		\caption{
			Framework for prototype-based cross-modal tracker incorporates the prototype generation algorithm. This algorithm consists of two key modules: the prototype evaluation module (PEM) and the prototype classification module (PCM). These modules work together to entails extraction and updating of multi-modal prototype in the prototype-based cross-modal tracker in various challenging scenarios.
			In the current example, PEM evaluates that the tracking result of the current frame is reliable, and decides to perform prototype extraction based on the tracking result. At the same time, PCM predicts the prototype category of the current frame as NIR. Consequently, the two modules decide to update the NIR samples by performing prototype updating.
		}
		\label{fig::overal_network2}
	\end{figure*}
	\subsubsection{Multi-Modal Prototype}
	Existing visual object tracking methods primarily focus on capturing temporal variations in target appearance. However, cross-modal tracking introduces additional challenges due to appearance variations across different modalities. As a result, existing methods \cite{yan2021learning, zhang2019learning} face difficulties in adapting to complex cross-modal scenarios.
	To overcome this limitation, we propose the innovative multi-modal prototype, which considers both temporal and modality variations in the target appearance. The multi-modal prototype is defined as follows:
	
	\begin{equation}
	P = (Z_{1}, Z_{RGB}, Z_{NIR})
	\end{equation}
	where $P$ represents the multi-modal prototype for the current frame, which represents target information by using multi-kind samples. $Z_{1}$ is a fixed sample extracted from the first frame, while $Z_{RGB}$ and $Z_{NIR}$ are representative samples from different modalities that are continuously updated during online tracking for the RGB and NIR modalities, respectively.
	
	The introduction of the multi-modal prototype provides several notable advantages for cross-modal tracking: 
	\begin{itemize}
		\item The fixed sample $Z_{1}$, derived from the first frame, provides a stable and accurate prototype representation of the target appearance. This ensures the availability of a reliable prototype representation throughout the entire tracking process, promoting precise target localization.
		\item The dynamically updated representative samples $Z_{RGB}$ and $Z_{NIR}$ offer a robust prototype representation of the target appearance. These samples possess the ability to adapt to variations in the target appearance over time and across different modalities, thereby enhancing the flexibility and adaptability of the cross-modal tracker.
		\item The multi-kind samples guarantees that a corresponding prototype representation is available for each modality, significantly improving the accuracy of the cross-modal tracker. Furthermore, the multi-kind samples can easily adapt to search regions of any modality, facilitating faster convergence and higher training accuracy of the tracker during the training phase and improving overall performance. In Figure \ref{fig::lossiou}, we provide a comparison of training loss and accuracy on the CMOTB dataset~\cite{li2022cross} between using the proposed multi-modal prototype that considers both temporal and modality variations and considering only temporal appearance variations in Stark~\cite{yan2021learning}.
	\end{itemize}
	
	By considering both temporal and modality variations in the target appearance, the proposed multi-modal prototype forms a robust target representation under temporal variation and modality switch, thereby enhancing the tracking robustness, accuracy, and adaptability in diverse cross-modal scenarios.
	
	\begin{figure*}[t]
		\centering
		\includegraphics[width=\linewidth]{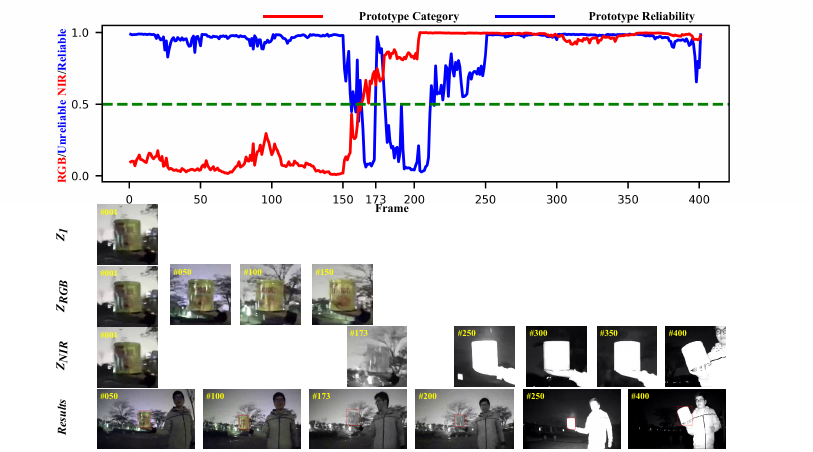}
		\caption{
			Visualization of prototype generation algorithm in a typical sequence.
		}
		\label{fig::plot}
	\end{figure*}
	
	\subsection{Prototype Generation}
	In this section, we will provide a detailed explanation of the prototype generation algorithm, which consists of three main components: the prototype classification module, the prototype evaluation module, and the integration of these two modules. Figure \ref{fig::overal_network2} illustrates the prototype-based cross-modal tracker that incorporates the prototype generation algorithm.
	\subsubsection{Overview}
	The prototype generation algorithm consists of the prototype classification module (PCM) and the prototype evaluation module (PEM). These modules work together to dynamically update multi-kind samples and ensure the representation of the multi-modal prototype in various challenging scenarios.
	
	The prototype classification module predicts the prototype category of each frame, determining whether it belongs to the RGB or NIR modality. This prediction guides prototype updating and the dynamic updates of the corresponding modality samples. The prototype evaluation module estimates the reliability of the tracking results for each frame, assessing whether the target is present within the search region. This assessment is crucial for deciding whether to perform prototype extraction based on the tracking result.
	
	To adapt to temporal changes in target appearance, we periodically update the multi-kind samples. Additionally, to adapt to modality switches in target appearance, we update the multi-kind samples of the corresponding modality when there is a change in the modality state. This approach ensures that the multi-modal prototype forms a robust target representation under temporal variation and modality switch.
	\subsubsection{Prototype Classification Module}
	In order to ensure the prototype representative in different challenges and achieve effective dynamic multi-kind samples updating, it is crucial to accurately determine the category of the current frame. In this paper, we leverage the availability of modality labels in the current cross-modal tracking dataset \cite{li2022cross} to facilitate the perception of the prototype category for each frame.
	
	Specifically, we introduce a prototype classification module to predict whether the current frame belongs to the RGB modality or the NIR modality. The design of the prototype classification module is straightforward, involving a simple binary classification task. It consists of three perceptron layers followed by a sigmoid activation function.
	\subsubsection{Prototype Evaluation Module}
	Merely predicting the prototype category of the current frame is insufficient for robust prototype updating. It is also crucial to determine whether it is appropriate to update the dynamic multi-kind samples based on the tracking results. For instance, erroneous prototype updating based on unreliable tracking results, such as occlusion or tracking drift, will have adverse effects on subsequent tracking. Moreover, incorrect prototype updating in subsequent frames after an initial reliable fixed sample will hinder the stability of the tracker, thereby impacting its performance. To address this challenge, we propose a prototype evaluation module to assess the reliability of the tracking results in the current frame.
	
	For simplicity, we predict whether the search region of the current frame contains the target. If the target is present in the search region, we consider the tracking results reliable and perform prototype extraction for prototype updating. Conversely, if the target is not present in the search region, we conclude that the tracking results are unreliable and refrain from prototype updating. This approach aligns with the setting in similar method, Stark \cite{yan2021learning}. The structure design of the prototype evaluation module is consistent with the prototype classification module, as both tasks involve simple binary classification, eliminating the need for overly complex network architectures.
	
	\subsubsection{Integration of Two Modules}
	To adapt to differences in target appearance over time and modality variations in cross-modal tracking, we jointly utilize the prototype classification module and prototype evaluation module for prototype generation. Specifically, we predict the prototype category $C_i$ of the search region in the $i$-th frame using the prototype classification module and record the prototype category $S$ of the previously updated dynamic multi-kind sample. If the prototype category $C_i$ of the $i$-th frame is inconsistent with the prototype category $S$ of the previously updated dynamic multi-kind sample, we consider it necessary to update the dynamic multi-kind sample for the current category to accommodate the modality variations.
	
	Additionally, we utilize the output of the prototype evaluation module $E$ to determine the reliability of the tracking results. If $E \geq 0.5$, indicating that the tracking results are reliable, we update the sample for the corresponding modality and update the prototype category $S$ of the previously updated dynamic multi-kind sample to match the prototype category $C_i$ of the current frame. Conversely, if $E < 0.5$, suggesting that the tracking results of the current frame are unreliable, we refrain from updating the dynamic multi-kind samples. The above process primarily considers the variation of the target appearance across modalities. In addition, we also take into account the temporal variation in the target appearance. Therefore, even if there is no modality variation, we perform dynamic multi-kind sample updates every 50 frames as long as the tracking results are reliable.
	
	We define the search region as $X$ and the fixed sample as $Z_1$. The initial two representative samples from different modalities, $Z_{RGB}$ and $Z_{NIR}$, are both set to $Z_1$. The algorithm iterates over each frame, predicting the prototype category $C_i$ using the prototype classification module and assessing the reliability of the tracking results $E$ using the prototype evaluation module. If necessary, the dynamic multi-kind samples $Z_{RGB}$ and $Z_{NIR}$ are updated based on the prototype category and reliability. Finally, the prototype category $S$ is updated accordingly. We also provide a visualization of the prototype generation algorithm in a typical sequence, as depicted in Fig.~\ref{fig::plot}.
	
	In summary, the prototype generation algorithm comprises the prototype classification module, prototype evaluation module, and their integration. These components work together to dynamically update multi-kind samples and ensure the prototype representative in different challenges. The algorithm leverages modality labels and evaluates the reliability of tracking results to guide prototype updating. By considering both temporal and modality variations, the prototype-based cross-modal tracker achieves robust target representation and enhances its performance in diverse cross-modal tracking scenarios.
	
	\subsection{Training and Inference}
	\subsubsection{Training}
	Our training approach is similar to Stark~\cite{yan2021learning}, which focuses on temporal appearance variations. However, our method extends beyond Stark by considering both modality and temporal appearance variations. Considering that joint training for localization and classification may lead to suboptimal results for both tasks~\cite{cheng2018revisiting, song2020revisiting}, we adopt a two-stage training algorithm to sequentially train the parameters of the tracking model and the parameters of the two modules.
	
	In the first stage, we train the parameters of the tracking model without training the parameters for the prototype classification module and prototype evaluation module. The training samples in this stage consist of search regions and multi-modal prototype. Specifically, the search regions contain the target, while the multi-modal prototype is randomly selected by choosing a fixed sample from the current sequence, followed by randomly selecting two representative samples from different modalities within the sequence. We employ the localization loss for training in this stage, where the localization loss is defined as:
	\begin{equation}
	L = \alpha L_{iou}(b_{gt},b_{predict}) + \beta L_{1}(b_{gt},b_{predict})
	\end{equation}
	
	
	The joint localization loss, which consists of the generalized IoU loss ($L_{iou}$)~\cite{rezatofighi2019generalized} and $l_1$ loss ($L_{1}$), is utilized to train the tracking model. It measures the discrepancy between the ground truth box ($b_{gt}$) and the predicted box ($b_{predict}$). In our experiments, we set the hyperparameters $\alpha$ and $\beta$ to 2 and 5, respectively, to modulate the impact of these losses. This approach of combining different loss functions has been empirically demonstrated to be effective in DETR~\cite{carion2020end} and Stark~\cite{yan2021learning}.
	
	In the second stage, we fix the parameters of the tracking model and train only the parameters of the prototype classification module and prototype evaluation module. The training samples consist of search regions that may or may not contain the target, fulfilling the requirement for confidence evaluation of the tracking results in the current frame. The multi-modal prototype sampling rule remains the same as in the first stage. Both modules are trained using binary cross-entropy loss, where the loss is defined as:
	
	\begin{equation}
	L_{pc} = y_{i}^{pc}\log{p_{i}^{pc}} + (1-y_{i}^{pc})\log({1-p_{i}^{pc}})
	\end{equation}
	\begin{equation}
	L_{pe} = y_{i}^{pe}\log{p_{i}^{pe}} + (1-y_{i}^{pe})\log({1-p_{i}^{pe}})
	\end{equation}
	\begin{equation}
	L = L_{pc} + L_{pe}
	\end{equation}
	
	The final loss is computed as the sum of the losses from both modules, where $y_{i}^{pc}$ and $y_{i}^{pe}$ represent the ground truth labels, and $p_{i}^{pc}$ and $p_{i}^{pe}$ denote the predicted probabilities for the prototype classification module and the prototype evaluation module, respectively. 
	
	It is worth noting that training a robust cross-modal tracker from scratch is challenging due to the relatively small size of the cross-modal object tracking dataset. Therefore, in the first stage, we import the pretrained parameters of tracking model, which are trained on larger datasets like LaSOT\cite{fan2019lasot}, GOT-10K\cite{huang2019got}, COCO\cite{lin2014microsoft}, and TrackingNet\cite{muller2018trackingnet}. We further fine-tune these parameters on the training set of the CMOTB dataset~\cite{li2022cross}, setting the learning rate to one-tenth of the pretrained stage. In the second stage, we train the parameters of the prototype classification module and prototype evaluation module for 50 epochs on the training set of the CMOTB dataset, while keeping the learning rate consistent with a similar method, Stark. The learning rate for these modules is set to 10e-4, and it is decayed by a factor of 10 after 40 epochs.
	
	\subsubsection{Inference}
	During the inference phase, the multi-modal prototype is initialized as the fix sample from the first frame. In each subsequent frame, we feed the search region and the multi-modal prototype into the tracking model to obtain the predicted target bounding box. We then use the prototype classification module and prototype evaluation module to predict the prototype category of the current frame and assess the reliability of the tracking results, respectively. If prototype updating is necessary, the dynamic representative sample for the corresponding modality is updated accordingly. 

	\begin{figure*}[!t]
		\centering
		\subfigure{\includegraphics[trim=0mm 0mm 0mm 0mm, clip, width=2.2in]{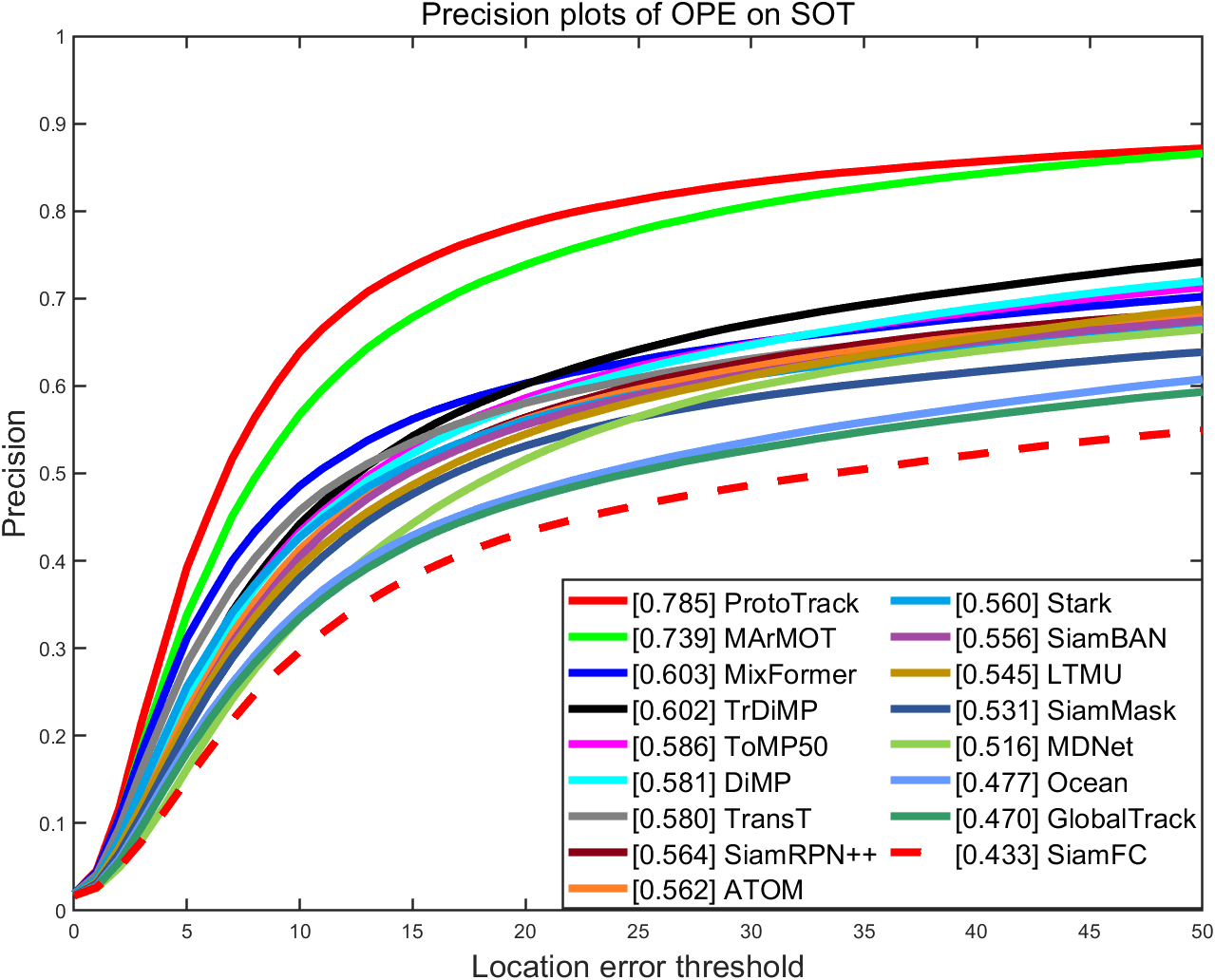}%
			\label{fig_first_case}}
		\hfil
		\subfigure{\includegraphics[trim=0mm 0mm 0mm 0mm, clip, width=2.2in]{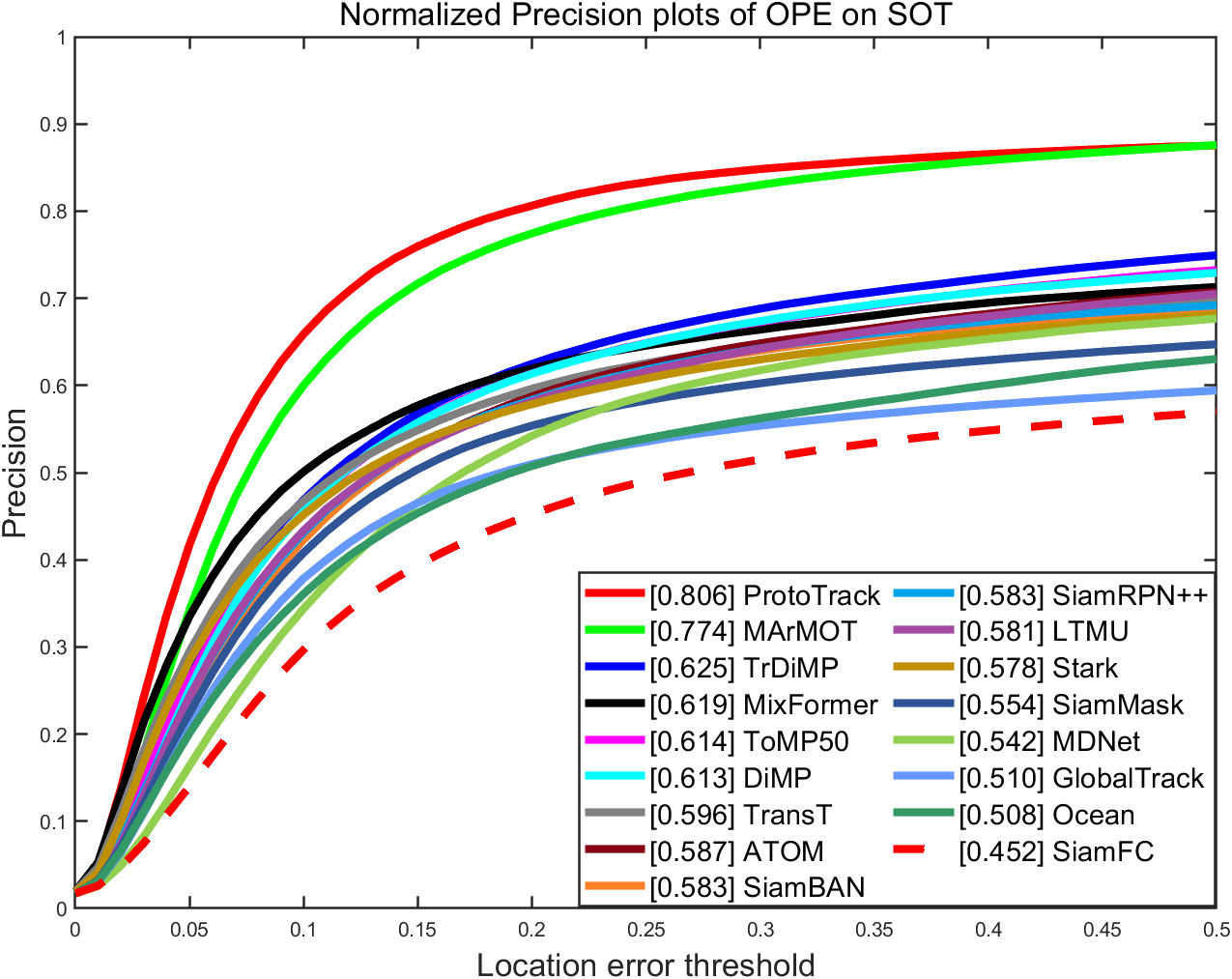}%
			\label{fig_second_case}}
		\hfil
		\subfigure{\includegraphics[trim=0mm 0mm 0mm 0mm, clip, width=2.2in]{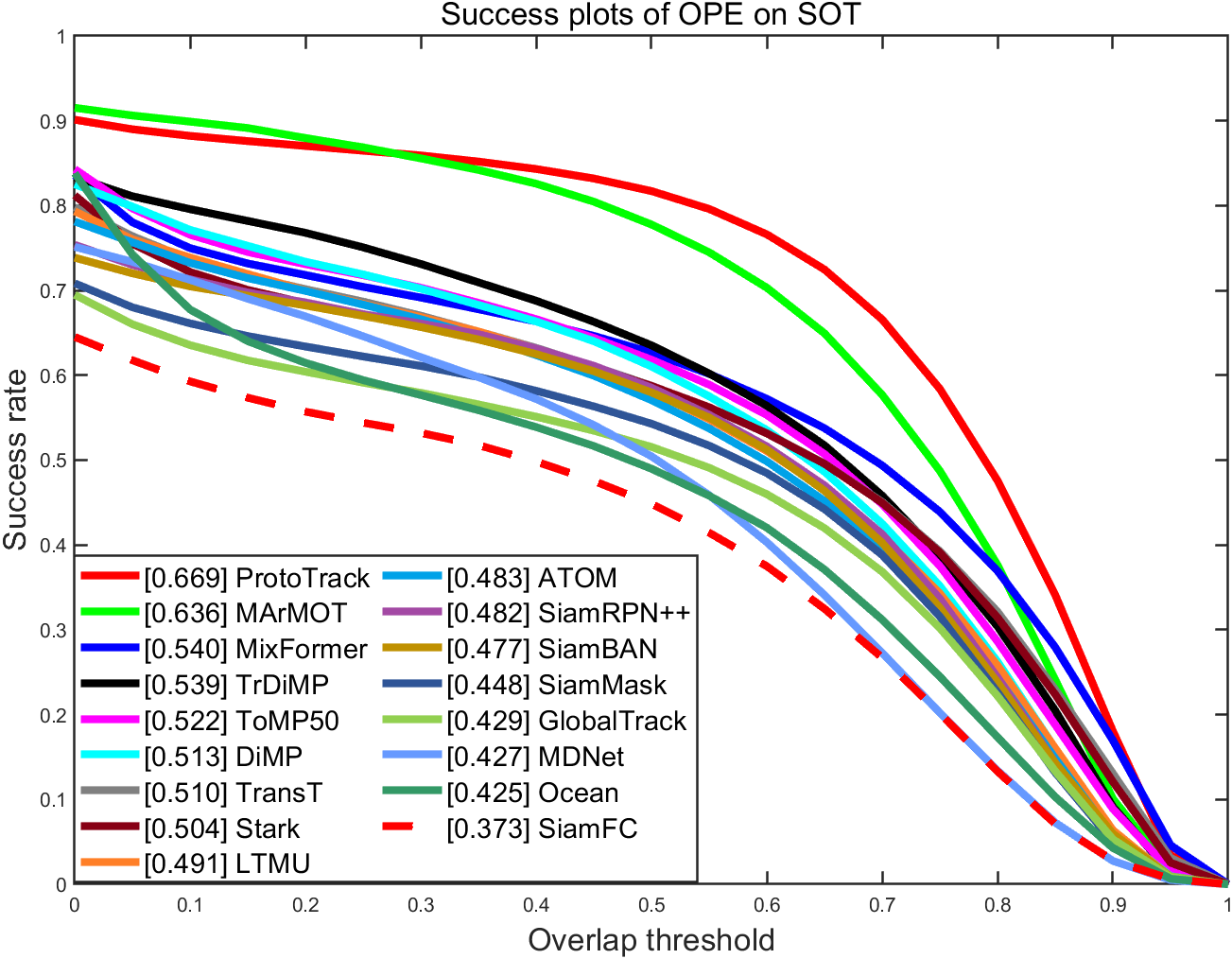}%
			\label{fig_third_case}}
		\caption{Comparisons on CMOTB testing set~\cite{li2022cross}.}
		\label{fig::overall_performance}
	\end{figure*}
	
	\subsection{Discussion}
	\subsubsection{Potentials of ProtoTrack}
	We believe that the prototype learning paradigm holds great potential for cross-modal object tracking, mainly manifested in the following aspects.
	\begin{itemize}
		\item {\bf Better adaptability compared to model update based tracking methods.} In cross-modal object tracking, the target appearance constantly changes over time and modality switches. Model update based methods require online collection of target appearance information and utilize gradient-based back propagation algorithms to update model parameters. However, the significant appearance variations caused by modality switches make it challenging to maintain stability in gradient updates~\cite{li2022cross}. Additionally, the implementation of gradient updates becomes difficult on edge devices~\cite{yan2021learning}, making it challenging to deploy such tracking algorithms.
		
		\item {\bf Better robustness compared to template based tracking methods.} For effective cross-modal object tracking, template based tracking methods require online accumulation of target templates to achieve stable tracking processes in response to target appearance changes. However, neither the initial template from the first frame nor the accumulated templates from last frames can maintain a robust target template to adapt to complex appearance variations, including temporal changes and modality switches. In contrast, the prototype learning paradigm can simultaneously learn prototype representations of targets under different modalities over time, enhancing the robustness of cross-modal trackers.
		
		\item {\bf Significant performance improvement.} The most critical challenge in cross-modal object tracking is adapting to significant appearance changes caused by modality switches. Prototype learning provides a promising solution by learning prototype representations under different modalities, achieving precise tracking performance in any modality.
	\end{itemize}
	\subsubsection{Differences of ProtoTrack From Template based Tracking Methods}
	The proposed ProtoTrack aims to efficiently incorporate prototype learning for cross-modal tracking. In contrast to existing few-shot learning methods that focus on learning representative representations for each category, our approach also considers learning prototype representations of the target that vary over time across different modalities. Similar to our tracking method, other approaches also employ template update strategies to adapt to target appearance variations. However, some of these methods rely on simple linear interpolation strategies~\cite{choi2018context, danelljan2016beyond, kiani2017learning} to update the target template. Unfortunately, these strategies overlook the severity of target appearance variations, such as occlusion and blurriness, where appearance variations are more pronounced. Furthermore, these methods tend to neglect the historical target appearance.
	In comparison to these methods, our multi-modal prototype retains fixed samples from the first frame, thereby enhancing the stability of the tracker. Additionally, our prototype generation algorithm incorporates a prototype evaluation module to ensure reliable prototype updating.
	
	Furthermore, there are also some tracking methods that combine the initial template from the first frame with the dynamic template to improve tracking stability. For example, Li et al. propose UpdateNet~\cite{zhang2019learning}, which utilizes the initial template, accumulated template from the last frame, and estimated tracking results of the current frame to predict the dynamic template for subsequent frames. Similarly, Stark~\cite{yan2021learning} uses the initial template and dynamically updated template to jointly estimate the results of the current frame. Although these strategies significantly improve the accuracy of template updating, they solely consider the temporal variation of target appearance and are ill-equipped to handle changes in target appearance due to modality switches.
	In contrast, our multi-modal prototype combines the fixed sample from the first frame with two dynamically updated representative samples from different modalities. This formation ensures a robust target representation under both temporal variation and modality switches. To further ensure the representativeness of the multi-modal prototype under different challenges, we design the prototype evaluation module and prototype classification module to control the process of prototype extraction and prototype updating. 
	
	\section{Experiments}
	Our tracker, ProtoTrack, is implemented in Python 3.6 and PyTorch 1.7, achieving a processing speed of approximately 24 frames per second (FPS) on a single A100 GPU. It is worth noting that this version of ProtoTrack is built upon Stark~\cite{yan2021learning}.
	In this section, we present a comprehensive evaluation of our tracker, ProtoTrack, through a series of extensive experiments. We first introduce the evaluation dataset and the metrics used for evaluation. Then, we compare the performance of our method against state-of-the-art trackers to verify its effectiveness. Additionally, we provide qualitative performance results to visually demonstrate the capabilities of our tracker. Finally, we conduct an analysis to validate the effectiveness and generalization ability of our proposed approach.
	
	\subsection{Dataset and Evaluation Metrics}
	\subsubsection{Dataset}
	To demonstrate the effectiveness of our proposed method, we conduct extensive experiments on the cross-modal object tracking benchmark CMOTB~\cite{li2022cross}. The CMOTB dataset comprises a total of 644 video sequences, with 325,000 frames in the training set and 153,000 frames in the testing set. These sequences are annotated with 11 challenging factors, including Scale Variation (SV), Background Clutter (BC), Aspect Ratio Change (ARC), Similar Object (SO), Fast Motion (FM), In-Plane Rotation (IPR), Out-of-View (OV), Partial Occlusion (PO), Modality Adaptation (MA), Full Occlusion (FO), and Motion Blurred (MB). This comprehensive list of challenging factors has been carefully curated and annotated to facilitate the evaluation of our tracker's performance under various attributes.
	
	In our experiments, we utilize the training set of CMOTB, comprising 430 sequences, to train our model. We carefully train our tracker on this diverse training set, enabling it to learn and adapt to a wide range of cross-modal tracking challenges. Following the training phase, we evaluate the performance of our tracker on the testing set of CMOTB, containing 214 sequences. This comprehensive evaluation allows us to evaluate the robustness and generalization capability of our approach across various challenging factors present in the CMOTB dataset.
	\begin{table*}[h]
		\centering
		\caption{The performance of various trackers on 11 challenging attributes using the CMOTB testing set is evaluated in terms of SR score. Additionally, a comparison of the tracking speeds of different trackers is presented. The \textcolor{red}{red} and \textcolor{blue}{blue} fonts are used to highlight the best and second best results, respectively.}
		{%
			\resizebox{1.\textwidth}{!}{
				\begin{tabular}{c|cccccccccccc|c}
					\hline
					\textbf{Trackers} & SV & BC & ARC & SO & FM & IPR & OV & PO & MA & FO & MB & ALL & FPS \\
					\hline
					\textbf{ProtoTrack} & \textcolor{red} {0.719} & \textcolor{red} {0.661} & \textcolor{red} {0.727} & \textcolor{red} {0.605} & \textcolor{blue} {0.625} & \textcolor{red} {0.668} & \textcolor{red} {0.712} & \textcolor{red} {0.636} & \textcolor{red} {0.713} & \textcolor{red} {0.633} & \textcolor{red} {0.603} & \textcolor{red} {0.669} & 24\\
					\textbf{MArMOT} & \textcolor{blue} {0.703} & \textcolor{blue}{0.626} & \textcolor{blue}{0.655} & \textcolor{blue}{0.593} & \textcolor{red}{0.637} & \textcolor{blue}{0.639} & \textcolor{blue}{0.693} & \textcolor{blue}{0.610} & \textcolor{blue}{0.646} & \textcolor{blue}{0.590} & \textcolor{blue}{0.586} & \textcolor{blue}{0.636} & 25\\
					\textbf{MixFormer}  & 0.612 & 0.523 & 0.603 & 0.503 & 0.508 & 0.536 & 0.614 & 0.513 & 0.555 & 0.469 & 0.456 & 0.540 & 75\\
					\textbf{TrDiMP}     & 0.577 & 0.523 & 0.540 & 0.536 & 0.574 & 0.518 & 0.595 & 0.521 & 0.508 & 0.514 & 0.475 & 0.539 & 18\\
					\textbf{ToMP}       & 0.541 & 0.487 & 0.511 & 0.495 & 0.540 & 0.533 & 0.587 & 0.504 & 0.511 & 0.504 & 0.482 & 0.522 & 28\\
					\textbf{Strak}      & 0.532 & 0.504 & 0.537 & 0.440 & 0.428 & 0.487 & 0.546 & 0.469 & 0.532 & 0.470 & 0.435 & 0.504 & 41\\
					\textbf{DiMP}       & 0.585 & 0.498 & 0.560 & 0.468 & 0.499 & 0.507 & 0.563 & 0.475 & 0.507 & 0.476 & 0.435 & 0.513 & 32\\
					\textbf{LTMU}       & 0.579 & 0.486 & 0.534 & 0.471 & 0.422 & 0.573 & 0.742 & 0.461 & 0.516 & 0.472 & 0.403 & 0.491 & 3\\
					\textbf{TransT}     & 0.590 & 0.512 & 0.556 & 0.485 & 0.489 & 0.568 & 0.651 & 0.481 & 0.506 & 0.490 & 0.429 & 0.510 & 26\\
					\textbf{SiamRPN++}  & 0.516 & 0.450 & 0.517 & 0.482 & 0.462 & 0.518 & 0.652 & 0.462 & 0.474 & 0.438 & 0.401 & 0.482 & 21\\
					\textbf{SiamBAN}    & 0.519 & 0.478 & 0.514 & 0.467 & 0.433 & 0.513 & 0.631 & 0.432 & 0.454 & 0.417 & 0.404 & 0.477 & 29\\
					\textbf{ATOM}       & 0.510 & 0.461 & 0.529 & 0.442 & 0.463 & 0.480 & 0.633 & 0.448 & 0.487 & 0.431 & 0.406 & 0.483 & 28\\
					\textbf{SiamMask}   & 0.501 & 0.434 & 0.492 & 0.444 & 0.473 & 0.513 & 0.612 & 0.413 & 0.440 & 0.376 & 0.355 & 0.448 & 41\\
					\textbf{GlobalTrack}& 0.408 & 0.439 & 0.448 & 0.364 & 0.318 & 0.473 & 0.602 & 0.388 & 0.459 & 0.428 & 0.395 & 0.429 & 1\\
					\textbf{MDNet}      & 0.453 & 0.395 & 0.437 & 0.419 & 0.421 & 0.435 & 0.540 & 0.407 & 0.376 & 0.388 & 0.385 & 0.427 & 1 \\
					\textbf{Ocean}      & 0.471 & 0.395 & 0.492 & 0.399 & 0.426 & 0.490 & 0.539 & 0.404 & 0.419 & 0.388 & 0.370 & 0.425 & 42 \\
					\textbf{SiamFC}     & 0.435 & 0.348 & 0.376 & 0.386 & 0.427 & 0.389 & 0.487 & 0.364 & 0.319 & 0.315 & 0.322 & 0.373 & 44 \\	
					\hline                 
				\end{tabular}%
			}
		}
		\centering
		\label{tb::results_challenges}
	\end{table*}
	\subsubsection{Evaluation Metrics}
	To evaluate the performance of our tracker, we utilize two evaluation metrics: precision rate (PR) and success rate (SR). PR measures the percentage of frames where the distance between the center point of the predicted bounding box and the ground truth is less than a predefined threshold. SR measures the percentage of frames where the overlap between the predicted and ground truth regions exceeds a certain threshold. These metrics provide a comprehensive assessment of the accuracy and robustness of our tracking method.
	
	In addition, we introduce a normalized precision rate (NPR) to account for the influence of target size on evaluation. Normalizing the PR metric allows us to evaluate the performance of our tracker across different target sizes, ensuring reliable evaluation performance.
	
	\textbf{NOTE} It is worth noting that the evaluation metrics we employ in this study differ from those mentioned in the paper of CMOTB ~\cite{li2022cross}. This discrepancy arises from identifying inaccuracies within the evaluation toolkit used in their paper. To address this issue, the authors have subsequently updated the results on their GitHub repository
	. For our experiments, we strictly adhere to the corrected evaluation toolkit provided by the authors, ensuring fair and accurate evaluation of our tracker's performance.
	
	\subsection{Comparison Results}
	We compare the performance of our proposed tracker, ProtoTrack, against 16 state-of-the-art trackers on the CMOTB dataset. The trackers we compare against include MDNet~\cite{nam2016learning}, SiamFC~\cite{bertinetto2016fully}, SiamMask~\cite{wang2019fast}, ToMP~\cite{mayer2022transforming}, GlobalTrack~\cite{huang2020globaltrack}, SiamRPN++~\cite{li2019siamrpn++}, ATOM~\cite{danelljan2019atom}, DiMP~\cite{bhat2019learning}, SiamBAN~\cite{chen2020siamese}, LTMU~\cite{dai2020high}, Ocean~\cite{zhang2020ocean}, TransT~\cite{chen2021transformer}, TrDiMP~\cite{wang2021transformer}, Stark~\cite{yan2021learning}, Mixformer~\cite{cui2022mixformer}, and MArMOT~\cite{li2022cross}.
	
	\subsubsection{Overall Performance}
	In Figure~\ref{fig::overall_performance}, we present the tracking performance using PR, NPR, and SR. The legends in the figure denote the representative scores for each metric.
	
	Our ProtoTrack achieves outstanding performance across all evaluation metrics on the CMOTB testing set, outperforming all other state-of-the-art trackers. Specifically, ProtoTrack achieves impressive scores of $78.5\%$, $80.6\%$, and $66.9\%$ for PR, NPR, and SR, respectively. These results demonstrate the effectiveness of our proposed method in accurately tracking targets in cross-modal scenarios. Our ProtoTrack also shows significant improvements over the baseline tracker, Stark~\cite{yan2021learning}, with gains of $22.5\%$, $22.8\%$, and $16.5\%$ in PR, NPR, and SR, respectively. When compared to the current best-performing tracker, MArMOT~\cite{li2022cross}, our ProtoTrack outperforms it with gains of $4.6\%$, $3.2\%$, and $3.3\%$ in PR, NPR, and SR, respectively.
	
	These results demonstrate the superior performance of our ProtoTrack, the robustness and adaptability of our ProtoTrack make it a promising solution for addressing the challenges of cross-modal object tracking.
	
	\subsubsection{Attribute-based Performance}
	To further demonstrate the effectiveness of our ProtoTrack, we conduct a comprehensive analysis of its performance across various challenging attributes commonly encountered by existing trackers. We evaluate our algorithm against 16 state-of-the-art trackers on 11 annotated attributes, as shown in Table ~\ref{tb::results_challenges}.
	
	The results clearly showcase the superior performance of our ProtoTrack across all annotated challenging attributes. Particularly noteworthy is the Modality Adaptation (MA) challenge, where the frame content exhibits high intensity due to imaging adaptation during modality switching. Our ProtoTrack achieves a remarkable improvement of $6.7\%$ in SR over the closest competitor, MArMOT. This result validates the effectiveness of our proposed method in effectively handling modality switches and adapting to different imaging characteristics.
	
	Furthermore, our performance excels in challenges such as Scale Variation (SV), Background Clutter (BC), Aspect Ratio Change (ARC), Similar Object (SO), In-Plane Rotation (IPR), Out-of-View (OV), Partial Occlusion (PO), Full Occlusion (FO), and Motion Blurred (MB). In each of these challenging scenarios, our ProtoTrack surpasses the capabilities of other state-of-the-art methods, demonstrating its potential for effectively addressing these specific challenges.
	
	These results highlight the robustness and versatility of our ProtoTrack in overcoming various difficulties encountered in real-world cross-model tracking scenarios. 
	
	\begin{figure*}[!thbp]
		\centering
		\includegraphics[width=1.\linewidth]{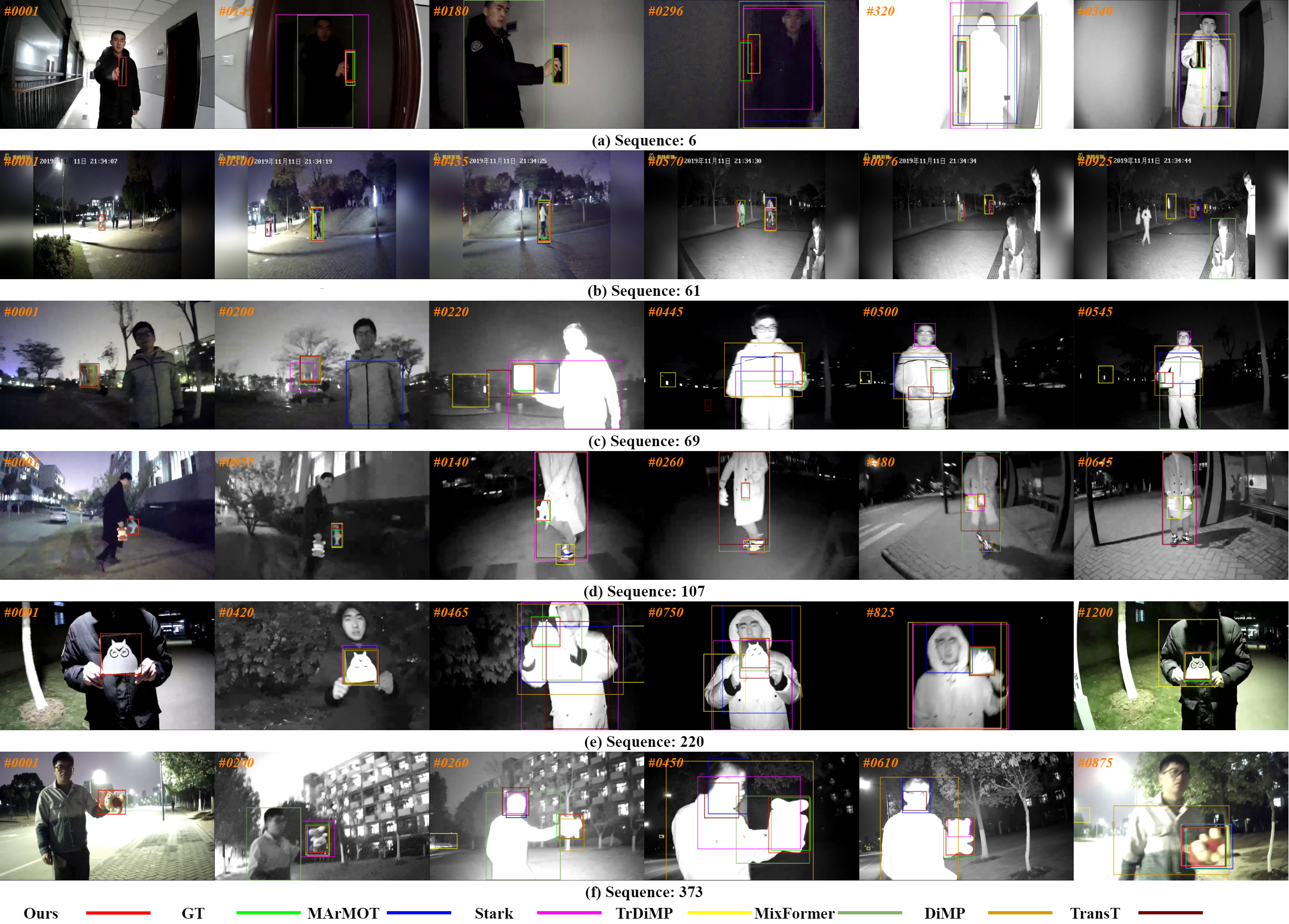}
		\caption{Visual comparison on six representative sequences.}
		\label{fig::visualization}
	\end{figure*}
	
	\subsubsection{Qualitative Performance}
	We provide qualitative performance results to visually demonstrate the capabilities of our ProtoTrack. In Figure \ref{fig::visualization}, we present visual comparisons between our ProtoTrack and several state-of-the-art trackers on six representative sequences. These sequences showcase challenging scenarios with appearance variations, providing a rigorous evaluation of the tracking performance.
	
	The results shown in Figure \ref{fig::visualization} highlight the exceptional capabilities of our ProtoTrack in handling appearance differences caused by cross-modal variations. Across all sequences, our method consistently outperforms the competing trackers after appearance variations, demonstrating the robustness and adaptability of our approach in tracking targets across different modalities.
	
	For example, in sequence 6, there is an abrupt modality switch at frame 320, which poses a significant challenge for most trackers. However, our ProtoTrack adeptly adjusts to the new modality and accurately locates the target. This exemplifies the superior performance of our approach in tackling modality variations. Additionally, in sequence 61, while many trackers can track the target successfully in the RGB modality, their performance significantly deteriorates in the NIR modality. In contrast, our method consistently maintains precise and reliable tracking in both modalities. This observation underscores the limitations of relying solely on an initial fixed sample derived from the RGB modality and highlights the critical importance of incorporating multi-modal prototype. Our ProtoTrack effectively leverages the information from multiple modalities, enhancing tracking performance and overcoming cross-modal challenges.
	
	These compelling findings validate the effectiveness of our ProtoTrack in addressing the inherent difficulties associated with cross-modal variations. The incorporation of multi-modal prototype allows our tracker to exploit the complementary information across modalities, leading to improved robustness and accuracy in cross-modal object tracking.
	
	\begin{table}[!t]
		\centering
		\caption{Comparison of several variants of our ProtoTrack on CMOTB test dataset to evaluate the impact of multi-modal prototype and prototype updating.}
		\resizebox{.325\textwidth}{!}{%
			\begin{tabular}{c|cccc}
				\hline
				Trackers & PR & NPR & SR \\
				\hline
				ProtoTrack-OS & 0.767 & 0.789 & 0.653\\
				ProtoTrack-TV & 0.760 & 0.783 & 0.652\\
				ProtoTrack-MV & 0.761 & 0.789 & 0.656\\
				\hline
				ProtoTrack & {\bf 0.785} & {\bf 0.806}  & {\bf 0.669}\\
				\hline              
			\end{tabular}%
		}
		\centering
		\label{tb::Ab12}
	\end{table}
	
	\subsection{Ablation Study}
	In this section, we conduct an ablation study to validate the effectiveness of the components of our ProtoTrack. We evaluate the impact and effectiveness of the proposed method, including the multi-modal prototype, prototype classification module, prototype evaluation module, and update interval.
	
	\subsubsection{Multi-Modal Prototype}
	The multi-modal prototype is designed to maintain two representative samples from different modalities that can adapt to the search region in any modality. To validate the effectiveness of this approach, we compare our ProtoTrack with a variant referred to as ProtoTrack-OS. In ProtoTrack-OS, only one dynamic sample is retained, which is similar to the approach used by Stark. And the dynamic sample in ProtoTrack-OS is updated when there is a modality switch or at a predefined update interval.
	
	The results of our comparative experiments, as presented in Table \ref{tb::Ab12}, clearly demonstrate a decline in performance for ProtoTrack-OS compared to the ProtoTrack. This decline serves as clear evidence of the effectiveness of the multi-modal prototype in ensuring robust tracking in any modality. The superior performance of the ProtoTrack underscores the significance of the multi-modal prototype in maintaining prototype representative in different challenges throughout the tracking process.
	
	\subsubsection{Prototype Generation Algorithm}
	Our prototype generation algorithm considers both target appearance differences over time and cross-modal variations. To validate the necessity of jointly considering these two types of variations, we conduct two additional comparative experiments named ProtoTrack-MV and ProtoTrack-TV. ProtoTrack-MV focuses solely on cross-modal variations, while ProtoTrack-TV focuses solely on time variations.
	
	We evaluate the performance of these two variants using PR, NPR, and SR as evaluation metrics. The experimental results, as shown in Table \ref{tb::Ab12}, demonstrate a decline in performance for both ProtoTrack-MV and ProtoTrack-TV compared to the ProtoTrack. For example, the PR for the ProtoTrack is $78.5\%$, while it drops to $76.1\%$ for ProtoTrack-MV and $76.0\%$ for ProtoTrack-TV. Similar drop trends are observed for NPR and SR.
	
	These declines in tracking accuracy provide strong evidence for the effectiveness of jointly considering target appearance differences over time and cross-modal variations in enhancing the robustness of the tracker. Our comparative analysis highlights the importance of accounting for both types of variations in developing a reliable cross-modal tracker and provides valuable insights for future research in this area.
	\begin{table}[!t]
		\centering
		\caption{Comparison of several variants of our ProtoTrack on CMOTB testing set to evaluate the impact of different update intervals.}
		\resizebox{.45\textwidth}{!}{%
			\begin{tabular}{c|cccc}
				\hline
				Update Interval & PR & NPR & SR & FPS\\
				\hline
				25 & 0.789 & 0.810 & 0.675 & 19 \\
				50 (Ours) & 0.785 & 0.806 & 0.669 & 24 \\
				100 & 0.777 & 0.800 & 0.665 & 27 \\
				200 & 0.763 & 0.787 & 0.652 & 28 \\
				500 & 0.764 & 0.791 & 0.658 & 30 \\
				\hline              
			\end{tabular}%
		}
		\centering
		\label{tb::Ab3}
	\end{table}
	\subsubsection{Update Interval}
	The update interval of the prototype updating plays a crucial role in effectively handling temporal variations in target appearance. A smaller interval allows the tracker to quickly adapt to changes in target appearance, potentially leading to improved performance. However, updating the prototype too frequently can negatively impact computational efficiency, which may hinder the tracking process. 
	To determine the optimal update interval, we evaluate different intervals, specifically 25, 50, 100, 200, and 500 frames, and analyze their impact on the performance of the tracker. The results, summarized in Table \ref{tb::Ab3}, include metrics such as PR, NPR, SR, and frames per second (FPS).
	
	After analyzing the results and considering the trade-off between accuracy and efficiency, we select an update interval of 50 frames as optimal. This interval strikes a balance between capturing and adapting to variations in target appearance while maintaining acceptable computational efficiency. This choice ensures that our tracker effectively handles temporal variations while achieving real-time tracking speed.
	
	The above ablation study confirms the effectiveness of the components in our ProtoTrack and provides insights into its contributions. The multi-modal prototype enables robust tracking in any modality, while jointly considering target appearance differences over time and cross-modal variations enhances the tracking performance. Additionally, selecting an appropriate update interval ensures a good balance between accuracy and efficiency in cross-modal object tracking.
	
	\begin{table}[!thbp]
		\centering
		\caption{Generalization ability of our prototype generation algorithm (PGA). Different trackers without and with PGA are evaluated using the CMOTB testing set.  * indicates the tracker is re-trained using the CMOTB training set.}
		\resizebox{.325\textwidth}{!}{%
			\begin{tabular}{c|cccc}
				\hline
				Trackers & PR & NPR & SR \\
				\hline
				OSTrack & 0.510 & 0.522 & 0.462\\
				OSTrack* & 0.750 & 0.772 & 0.644\\
				OSTrack+PGA & 0.800 & 0.817  & 0.677\\
				\hline
				Stark & 0.560 & 0.578 & 0.504\\
				Stark* & 0.728 & 0.752 & 0.628\\
				Stark+PGA & 0.785 & 0.806  & 0.669\\
				\hline              
			\end{tabular}%
		}
		\centering
		\label{tb::Ab4}
	\end{table}
	
	\subsection{Generalization Ability}
	We observe that our multi-modal prototype and prototype generation algorithm can be easily integrated into a wide range of tracking algorithms. To demonstrate the remarkable generalization capabilities of our algorithm, we have incorporated it not only into our baseline method, Stark~\cite{yan2021learning}, but also into OSTrack~\cite{ye2022joint}, which is considered one of the most influential transformer-based trackers in the tracking field.
	
	In Table \ref{tb::Ab4}, we provide a comprehensive comparison of the tracking performance on the CMOTB testing set, demonstrating the results achieved with and without the prototype generation algorithm. The results clearly indicate a consistent improvement in tracking accuracy when utilizing our proposed prototype generation algorithm on different tracking frameworks. These findings effectively validate the generalization ability and effectiveness of our algorithm.
	
	\section{Conclusion}
	In this paper, we propose the prototype-based cross-modal tracker (ProtoTrack), a robust cross-modal object tracking method. Our approach addresses the challenge in adapting to significant target appearance variations in the presence of modality switch.
	The ProtoTrack incorporates a multi-modal prototype to represent target information by multi-kind samples to forms a robust target representation under temporal variation and modality switch, including a fixed sample from the first frame and two representative samples from different modalities. To ensure the prototype representative in different challenges, we develop a prototype evaluation module to determine whether to perform prototype extraction based on the tracking result, and a prototype classification module to facilitate the dynamic prototype updating of the associated modality samples. Experimental results on public dataset and two tracking frameworks demonstrate the superiority and generalization ability of our method.
	\bibliographystyle{IEEEtran}
	\bibliography{mafnet}
	
\end{document}